# Query-Aware Source-Risk Triage for Retrieval-Augmented Generation

Kainan Zhou
*Google LLC*
Mountain View, USA
zhoumark@google.com

Gangzhen Qian
*Google LLC*
Mountain View, USA
irisqian@google.com

Chuhong Xu
*Sony Corporate of America*
San Jose, USA
chuhong.xu@sony.com

Lu Yi
*Google LLC*
Mountain View, USA
annabelyi@google.com

***Abstract*—Retrieval-augmented generation (RAG) pipelines may omit a source's material relationship to the query. We study a pre-generation triage layer that treats this relationship as query dependent. The method routes canonical query families for enhanced review and assigns retrieved pages to pass, contextualize, exclude, or review. It combines a four-dimension page score, rank-discounted family aggregation, intent-preserving query mutations, and a family-held-out router. A single-coded pilot of 200 real URLs supplies provisional calibration anchors; a 20,000-row scenario with synthetic domain identifiers supports controlled workload analysis. An oracle page gate defines a risk–coverage target for a future learned classifier. The evaluation shows why page-level frequency cannot substitute for family-level exposure and quantifies how calibration changes scenario activation. Annotation reliability remains unmeasured, and synthetic rankings omit real retrieval dynamics. The result is an auditable triage method and validation plan, not an estimate of deployed review workload, live-Web prevalence, or downstream answer-quality gains.**

***Keywords*—*retrieval-augmented generation, source provenance, query-aware triage, query clustering, third-party bias***

## I. Introduction

Retrieval-augmented generation (RAG) combines search, ranking, and synthesis. Verifiability audits distinguish citation completeness from whether each citation supports its associated statement [1]. Those tests are necessary, but they do not cover a separate source-side failure. A vendor comparison may report accurate specifications, an affiliate review may contain useful measurements, and an advocacy group may cite primary records. Trouble arises when synthesis removes this material relationship and presents evidence from an interested source as independent.

End-to-end evaluation must also separate retrieval, citation placement, and answer generation; ALCE evaluates all three and shows that fluent output can conceal upstream failure [2]. Source relationships arise in both ordinary and adversarial retrieval, but an interested source is not necessarily false or malicious. This distinction motivates examining provenance after retrieval and before generation.

Our two-stage framework first routes query families for enhanced provenance review and then defines query-page policy actions: pass, contextualize, exclude, or review, without permanently labeling publishers.

The evaluation uses two complementary components. A manual audit of 200 real URLs supplies page-score anchors and records access failures. A frozen scenario of 400 constructed query families and 20,000 query-page rows uses those anchors to test aggregation, threshold workload, controlled mutations, and family-held-out routing. Because the manual means calibrate the scenario, agreement between the two is by construction rather than validation. We therefore ask three bounded questions: how should query-page relationship risk be represented, how does manual anchoring affect family activation, and what risk–coverage target should a later page classifier meet?

The resulting prototype contributes a query-page relationship score with rank-aware family activation, a family-held-out router test for controlled query reformulations, and an oracle risk–coverage target for a future learned page classifier. Fig. 1 summarizes the workflow. The present study does not include matched downstream answer generation.

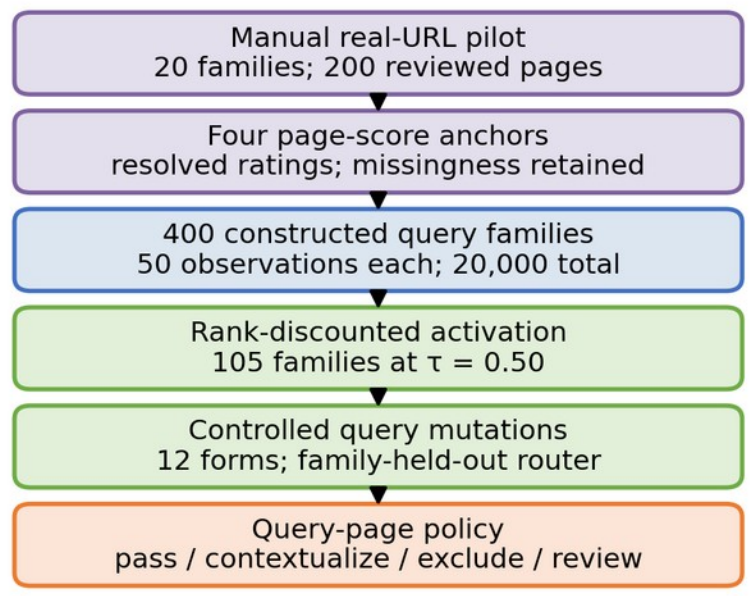

Fig. 1. Query-aware triage workflow. Manual page scores calibrate the constructed scenario; family routing precedes query-page policy actions.

## II. Related Work and Scope

### A. RAG and Citation Evaluation

Citation quality requires more than reference presence: SIDE identifies claims needing citations and retrieves stronger sources for Wikipedia statements [3]. This motivates checking whether each retrieved source supports the attached factual claim.

Our scope is complementary to answer-level hallucination detection: it asks whether a supported passage comes from a materially interested source, while keeping factual support and provenance as separate review dimensions.

### B. Source-Side Influence and Disclosure

Source-side influence can arise through optimization, injected documents, affiliate infrastructure, incomplete disclosure, coordination, or undisclosed sponsorship. These

mechanisms differ in intent, so our signal remains query dependent and does not prove deception.

### C. Unit and Claim Boundary

The unit of analysis is a query-page pair (q, u), not a person, publisher, or domain. Its score is a query-specific review-priority index, not the probability that u is false, deceptive, coordinated, or malicious. We use four evaluation strata: product recommendation (PROD), software or service comparison (SOFT), political or electoral information (POL), and your-money-or-your-life topics (YMYL), including health, finance, or local services. These strata organize evaluation, not Web prevalence.

### D. Threat Model and Operating Objective

The method addresses retrieval sets where relationship evidence may be visible on a page, in a redirect chain, or in public ownership records, yet absent from the excerpt delivered to a generator. A source may enter through ordinary ranking rather than adversarial insertion. The failure is contextual: the generator receives a useful passage but lacks information needed to interpret the source's stake. We assume neither control of the retriever nor access to private ranking features. The objective is provenance review before synthesis while preserving sufficient evidence after policy actions. Deliberate poisoning is one case, but the same interface covers sponsored, affiliated, or advocacy material in ordinary retrieval.

## III. Methodology

### A. Manual Calibration Corpus

Each query-page pair receives four ordinal ratings in {0, 0.25, 0.50, 0.75, 1}: query-relevant incentive or material relationship $I$, directional advocacy $A$, disclosure opacity $O$, and apparent independence $P$. We combine them as

$$S_{\text{page}}(q,u)=(I\,A\,O\,P)^{1/4}. \tag{1}$$

The geometric mean is deliberately conjunctive: any zero yields a zero composite. The score is an operational review-priority index; a low value does not certify safety or independence, nor is it a probability of falsehood, deception, or malicious intent.

The manual workbook contains 20 English audit queries, five per vertical, with ten recorded URLs per query. Two reviewers divided the 200 pages equally and recorded source role, evidence notes, access status, and the four dimensions. No pages were independently double-rated, so reliability and inter-rater agreement remain unmeasured. This is a single-coded pilot, not a human gold standard. Section VII specifies an overlap audit. Reviewer-entered titles only aid navigation and are excluded from evidence.

Fourteen pages remain unresolved: six were unavailable, four led to the wrong page, and four were partial or paywalled. They remain missing rather than being imputed as safe. Most fall in one reviewer's assignment, but the disjoint allocation prevents reviewer and access effects from being separated.

### B. Manual-Anchored Scenario Generator

The scenario uses a frozen panel of 400 constructed canonical queries, 100 per vertical. Each query has 50 ranked synthetic source observations, yielding 20,000 query-page rows. Synthetic domain identifiers do not correspond to real publishers; no search engine, page crawl, model call, or human rating produced these rows. The generator includes topic, intent, domain, rank, and noise terms, with fixed rank component 0.22 $\exp(-(r-1)/3.2)$ and fixed random seed 20260716.

We fit one additive intercept $\delta_v$ per vertical. For raw score $S_{\text{raw}}$ and resolved manual mean $\mu_v$, $\delta_v$ solves

$$\mathrm{E}\left[\operatorname{clip}\left(S_{\text{raw}}+\delta_v,0,1\right)\mid v\right]=\mu_v. \tag{2}$$

Only four intercepts are fitted; topic, query, rank, domain, and noise structure remain frozen. Raw and anchored scores are both retained. Each intercept changes only its vertical baseline, so subsequent activation changes arise from the declared calibration step. The goal is to prevent arbitrary vertical offsets from dominating workload analysis, not to validate the generator against the records used for calibration.

### C. Rank-Discounted Family Score

For a query with $n_q$ ranked pages, ranks receive weights proportional to $1/\log_2(r+1)$, normalized within that query. Long-context experiments show that evidence position affects model use, which motivates rank-discounted exposure while page policy remains local [4]. The family score is

$$S_{\text{query}}(q)=\sum_{r=1}^{n_q} w_r\,S_{\text{page}}(q,u_r),\ \sum_{r=1}^{n_q} w_r=1. \tag{3}$$

Here $n_q$ = 10 for the manual audit and $n_q$ = 50 for the constructed scenario. A family activates when $S_{\text{query}}(q) \geq \tau$ and at least two constituent pages also satisfy $S_{\text{page}}(q,u_r) \geq \tau$. The main analysis uses $\tau$ = 0.50 and repeats the same rule from 0.40 to 0.60. The two-page guard prevents a single extreme page from determining family activation. Page-level policy decisions remain local to each page within the originating query.

### D. Controlled Mutations and Router

Each canonical query produces 12 textual forms: the original wording and 11 intent-preserving mutations spanning seven categories: comparison, evidence-first, skeptical, disclosure-aware, constraint-aware, alternative-seeking, and independent-source. All forms from one canonical query remain in the same train, development, or test partition. The 60/20/20 family split therefore prevents near-duplicate formulations from crossing partitions.

We compare exact-string canonical matching, a development-tuned risk lexicon, word term frequency–inverse document frequency (TF–IDF) logistic regression, and a word-plus-character TF–IDF router. The prediction target is the anchored scenario activation rule rather than a human query label. Results on 80 held-out families therefore measure recovery of controlled reformulations under that rule, not accuracy on natural or deployed queries.

## E. Oracle Page Gate

The prototype evaluates the page-policy interface with an oracle observing the manual page score. At threshold τ, resolved pages with $S_{\text{page}}(q,u) \geq \tau$ are withheld; unresolved pages go to review. RAGAS separates context relevance from faithfulness, motivating distinct retention and residual-risk measures [5]. We also require at least three retained pages as a fixed evidence-sufficiency proxy. The resulting curve defines a risk–coverage target for learned page models. Table I summarizes each component and its claim boundary.

The interface contains four actions: pass, contextualize, exclude, and review. Pass retains a page without certifying it unbiased; contextualize retains evidence while disclosing the relationship; exclude withholds evidence when neutral synthesis is inappropriate; and review handles missing or insufficient provenance. The oracle implements pass, exclude, and review; contextualization requires answer generation and remains outside the current evaluation.

TABLE I. EVIDENCE ROLES IN THE EVALUATED PROTOTYPE

| Component | Records | Evaluated purpose | Unsupported inference |
|---|---|---|---|
| Manual audit | 200 URLs | Calibration and missingness | Annotation reliability or prevalence |
| Scenario | 20,000 rows | Workload and threshold stress | Live-Web behavior |
| Query router | 4,800 forms | Controlled mutation recovery | Deployed-query accuracy |
| Oracle gate | 200 URLs | Risk–coverage target | Classifier accuracy or answer effect |

# IV. RESULTS

## A. Manual Page and Query Units

Among the 200 manual records, 186 have resolved scores, and 72 of those 186 (38.7%) meet the 0.50 threshold. Mean scores are 0.477 for PROD, 0.418 for SOFT, 0.291 for POL, and 0.248 for YMYL. PROD and SOFT each have a 55.3% high-page share; the corresponding shares are 23.9% for POL and 19.6% for YMYL. The cluster-bootstrap intervals in Fig. 2 overlap, and each vertical contains only five audit queries. These values therefore serve as calibration anchors rather than population estimates.

Fig. 3 illustrates the effect of manual anchoring on activated-family counts at the main threshold. Rank-discounted aggregation yields a different family-level pattern. Fig. 4 presents the highest-scoring anchored constructed query families. In the manual audit, by contrast, only PROD-003 (0.614) and PROD-001 (0.560) activate at τ = 0.50. No SOFT audit query activates, even though SOFT has the same page-level high-score share as PROD. This contrast illustrates why family activation depends on aggregate exposure rather than raw page counts alone across query families.

## B. Calibration Effect and Workload

Table II reports the manual-anchored scenario statistics across verticals. Fig. 5 illustrates the calibration shifts and their effect on family activation. The fitted intercepts are −0.025 for PROD, −0.049 for SOFT, −0.056 for POL, and −0.012 for YMYL. All four lower the raw scenario scores because the unadjusted means exceed the manual anchors. At τ = 0.50, calibration reduces activated families from 119 to 105: 45 PROD, 33 SOFT, 18 POL, and 9 YMYL. SOFT shows the largest change, falling from 41 to 33 activated families.

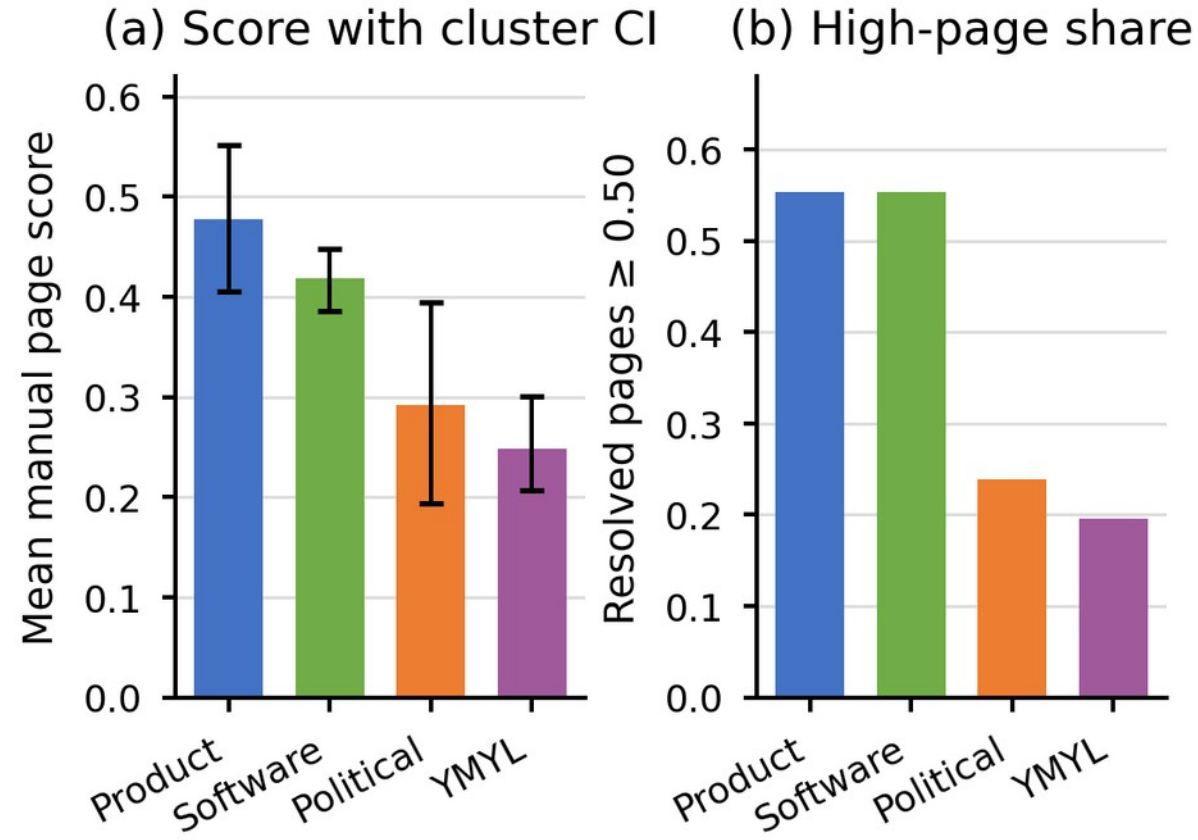

Fig. 2. Single-coded manual real-URL pilot. Error bars are 95% query-cluster bootstrap intervals from five query families per vertical.

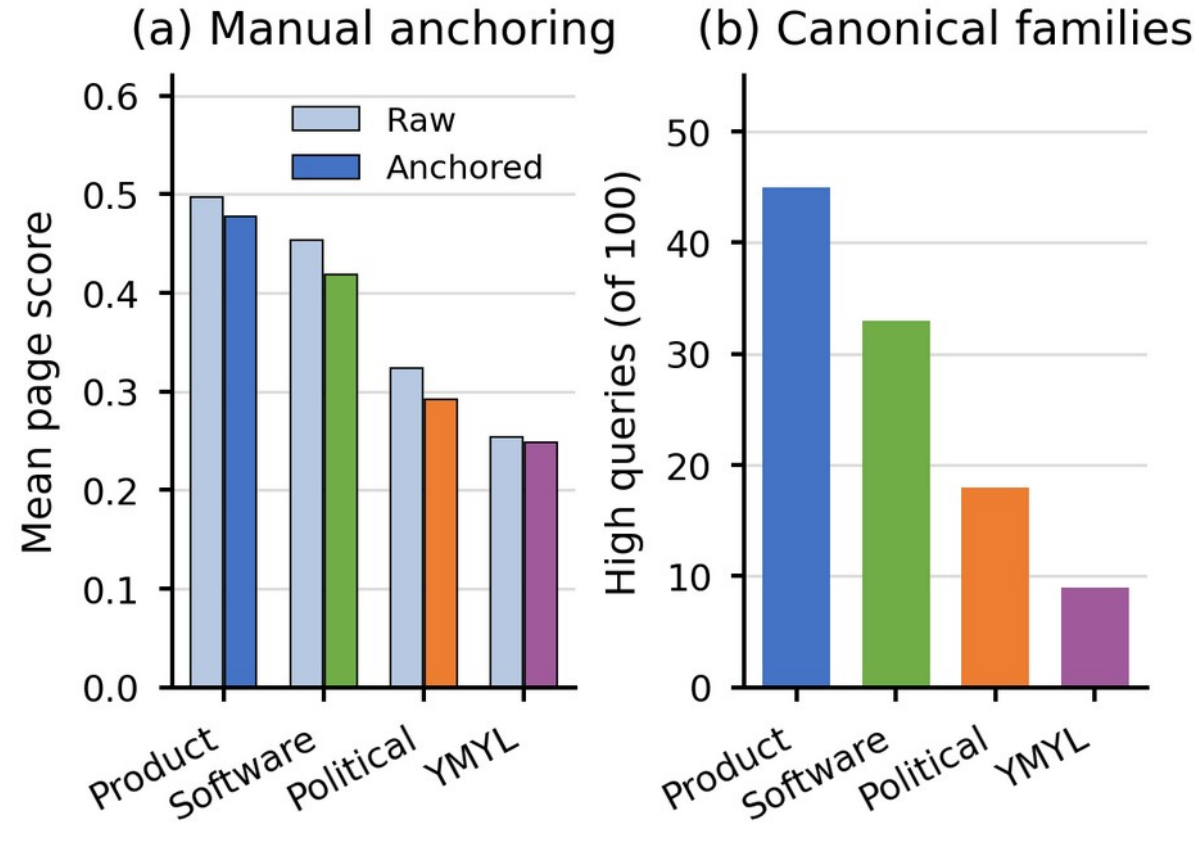

Fig. 3. Manual anchoring and activated-family counts at τ = 0.50.

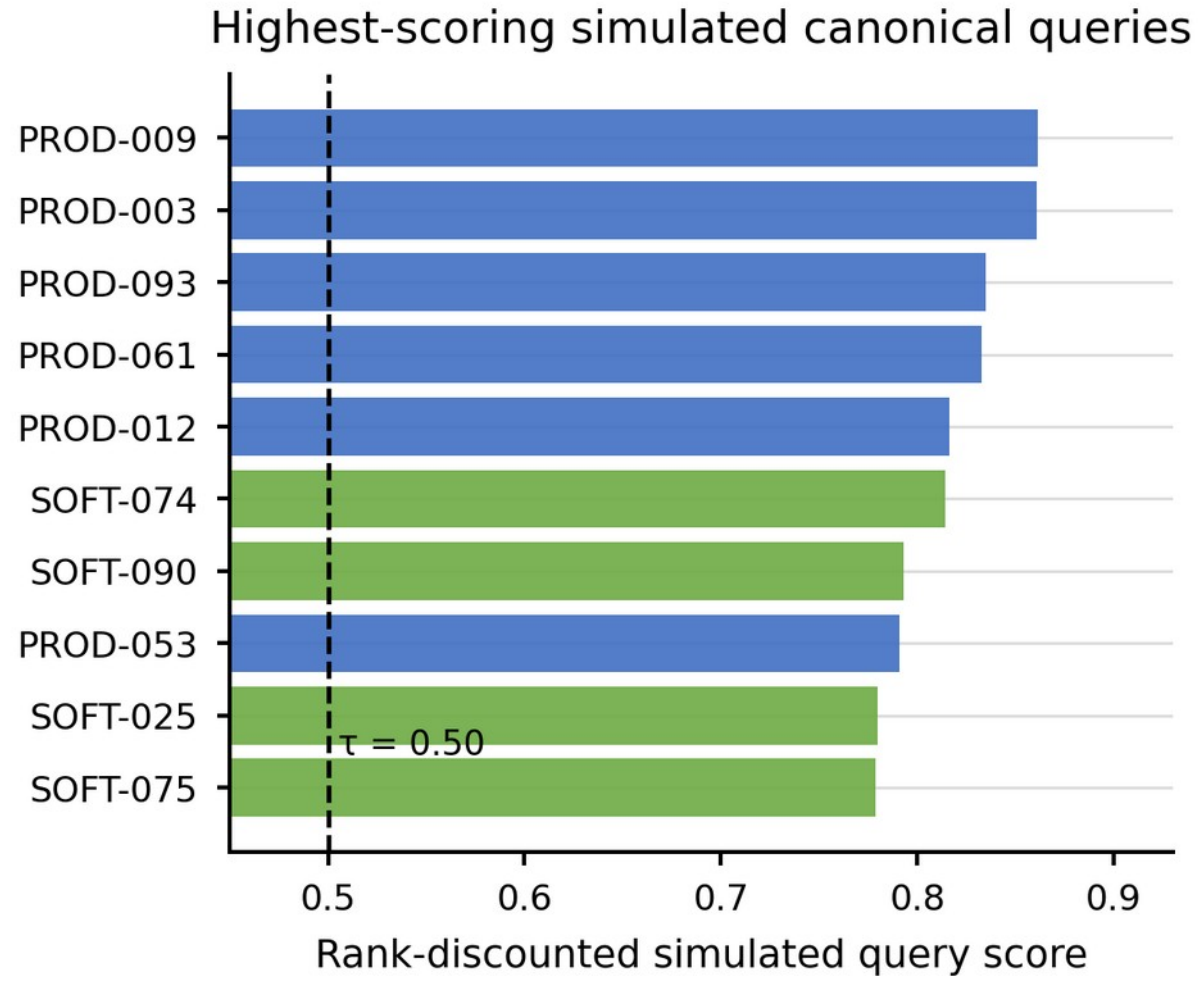

Fig. 4. Highest-scoring anchored canonical families.

TABLE II. MANUAL-ANCHORED SCENARIO BY VERTICAL

| Vertical | Raw mean | $\delta_v$ | Anchored mean | Activated |
|---|---|---|---|---|
| PROD | 0.497 | −0.025 | 0.477 | 45/100 |
| SOFT | 0.453 | −0.049 | 0.418 | 33/100 |
| POL | 0.324 | −0.056 | 0.291 | 18/100 |
| YMYL | 0.253 | −0.012 | 0.248 | 9/100 |

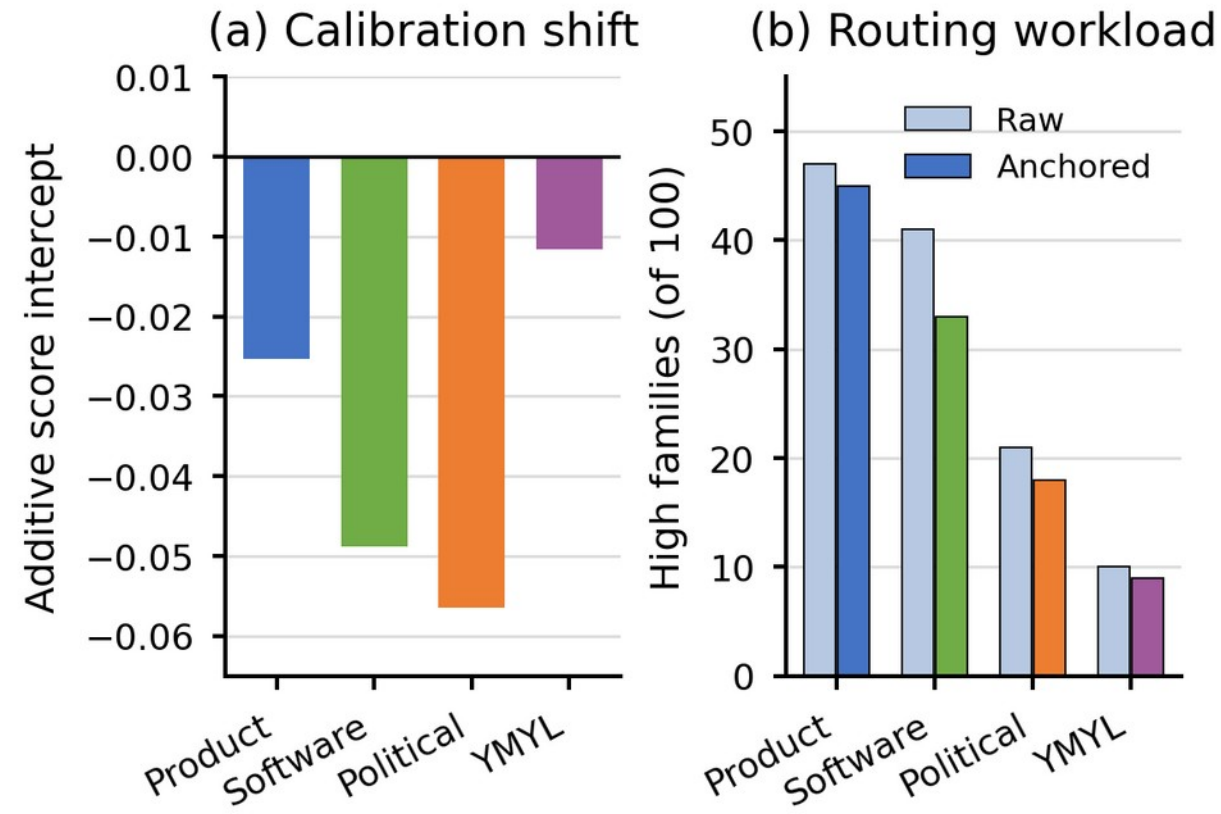

Fig. 5. Calibration shifts and their effect on family activation. Anchored means equal manual means by construction.

Table III reports threshold sensitivity: 163, 128, 105, 90, and 67 families activate at 0.40, 0.45, 0.50, 0.55, and 0.60, respectively. At 0.50, activation is 105/400 (26.3%) in a panel with equal vertical shares and 50 synthetic ranks per query. This is conditional workload, not deployed review load or live-query prevalence. Query mix, ranking rules, publisher concentration, and retrieval depth may change activation; these effects require validation with actual retrieved sources.

TABLE III. ANCHORED THRESHOLD SENSITIVITY

| $\tau$ | Activated | Audit activated | Evidence retained |
|---|---|---|---|
| 0.40 | 163/400 | 7/20 | 0.459 |
| 0.45 | 128/400 | 4/20 | – |
| 0.50 | 105/400 | 2/20 | 0.617 |
| 0.55 | 90/400 | 2/20 | – |
| 0.60 | 67/400 | 1/20 | 0.795 |

### C. Family-Held-Out Router

As expected under family-held-out evaluation, exact canonical matching has zero recall on unseen mutations. The risk lexicon reaches 0.401 precision, 0.625 recall, and 0.489 F1. Word TF–IDF logistic regression reaches 0.521 F1. The word-plus-character TF–IDF model reaches 0.546 precision, 0.492 recall, 0.518 F1, and 0.748 accuracy. The two learned routers therefore have similar F1 at this scale, while the word-plus-character model favors precision over recall.

These results support a limited engineering claim: simple text models recover part of the scenario-derived activation rule under unseen controlled reformulations. A missed activation can bypass provenance analysis, whereas a false activation increases review workload. Future tuning should therefore report both activation recall and activation load. Validation still requires independent labels across natural query variants.

### D. Oracle Risk–Coverage Target

Fig. 6 shows the oracle page-gate risk–coverage target under different thresholds. Before gating, mean rank-discounted manual score mass is 0.346. At $\tau = 0.50$, the oracle retains 61.7% of discounted evidence and leaves residual score mass 0.099. Nineteen of 20 queries meet the three-page evidence-sufficiency proxy, with 5.7 retained pages on average. At 0.40, evidence retention drops to 45.9%, and 17 queries meet the proxy. At 0.60, retention rises to 79.5%, all 20 queries meet the proxy, and residual score mass increases to 0.195.

The reduction at $\tau = 0.50$ is mechanically coupled to the score observed by the oracle. ARES likewise separates context relevance, answer faithfulness, and answer relevance [6]. Our result is neither classifier accuracy nor an answer-quality effect; it defines a retained-evidence level for comparing a future page model against residual manually scored mass.

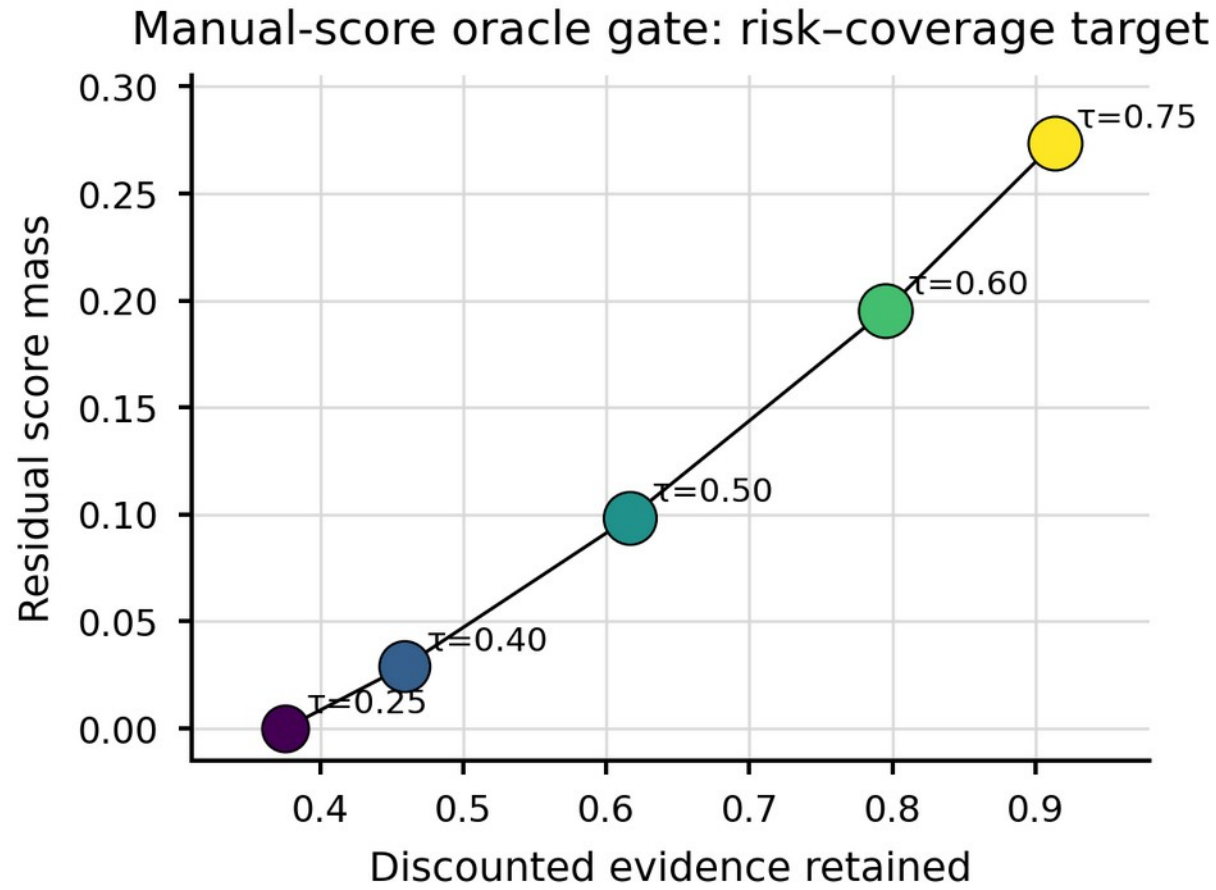

Fig. 6. Oracle page-gate target. Marker size gives the share of queries satisfying the retention proxy of three pages.

## V. TRIAGE AND EVALUATION CONTRACT

### A. Runtime Path

Seven reproducible modules form the evaluated prototype: a frozen scenario-query panel, ranked scenario generation, manual vertical calibration, rank-discounted aggregation, controlled mutation, a family-held-out router, and oracle policy evaluation. Because content edits can alter visibility in generative engines [7], exported rows retain page provenance, origin, family, rank, raw and calibrated scores, and threshold decisions for regenerating workload and router results.

At runtime, the trained family router receives query text and returns an activation decision. Only activated queries request deeper query-page provenance analysis. Retrieved documents can steer generated answers under PoisonedRAG [8], so provenance review remains a runtime concern rather than a publisher label. Unactivated queries still undergo ordinary relevance and citation-support checks; routing prioritizes provenance review while leaving those checks intact.

### B. Page Policy

A learned model should replace the oracle only if it uses query-page evidence rather than domain reputation alone.

Relevant inputs include ownership, sponsorship, redirects, advocacy framing, evaluated passages, and uncertainty; affiliate-abuse measurements show why redirects and tracking are evidence-bearing features [9]. Evaluation should hold out canonical families and domains so memorized publishers or disclosure templates cannot masquerade as generalization.

Filtering is only one policy action; disclosure studies document inconsistent affiliate labeling on social platforms [10]. A retained set may overrepresent one publisher or omit unique evidence, while a high-scoring page may still contain supported information. Selection should balance risk, relevance, support, and publisher diversity under an evidence-sufficiency constraint. Contextualization can preserve useful material with relationship disclosure beside each affected claim.

### C. Downstream Test

An answer-level effectiveness claim requires matched generation from the same archived candidate set under control, filter, and contextualize conditions. Because AdIntuition infers endorsement disclosure from content [11], the primary endpoint is whether contextualization preserves relationship information near the affected claim. Model version, prompt, decoding, source order, and evidence budget should be fixed, and evaluators should remain blind to condition. Support, answerability, abstention, diversity, hallucination, and latency provide complementary measures.

No matched answers or answer ratings are included in the present study. The downstream experiment is specified so that a later implementation can test the interface without attributing answer-level improvement to the current evidence.

The same archived candidates should feed every condition. Rerunning retrieval after filtering would confound the policy action with changes in rank or page availability. Each answer record should preserve candidate order, retained passages, exclusions, disclosures, abstention, and the final evidence budget. This design allows failures to be localized to retrieval, page assessment, selection, or generation rather than assigning every error to the triage layer.

## VI. Discussion and Limitations

The page- and family-level results support keeping activation and page treatment separate. Coordination patterns used to detect political astroturfing also show why publisher-level labels can be too coarse [12]. PROD and SOFT have the same high-page share in the manual audit, yet only PROD contains activated audit queries. In the scenario, calibration changes workload across verticals, while family-held-out routing provides a stronger test than exact matching.

The oracle curve exposes the central trade-off: reducing residual relationship-risk mass also removes evidence. Affiliate fraud can remain invisible in a quoted passage [13], so the page score determines review priority rather than deception or factuality. Evaluation should pair risk reduction with retention, the three-page evidence-sufficiency proxy, and publisher diversity. Missing evidence goes to review rather than being treated as safe.

Manual anchoring transfers four vertical means, not the joint distribution of real queries, publishers, and ranks. Synthetic identifiers and fixed rank terms omit retrieval-dependent publisher concentration, duplicate sources, temporal changes, and access failures. These omissions can affect activation and retained evidence. Thus, 20,000 rows support controlled workload analysis, not independent validation. The 26.3% activation is conditional on the generator; Section VII specifies an archived-retrieval check to assess transfer to actual retrieval.

RAGTruth labels unsupported RAG output at word level [14], whereas our provenance score asks whether a supported passage comes from a materially related source. Relevance, factual support, and provenance therefore remain separate: each governs passage utility, claim support, or relationship disclosure. This distinction matters especially when exclusion would remove unique evidence.

The single-coded pilot establishes neither annotation reliability nor codebook replicability. The query-cluster intervals in Fig. 2 reflect variation across sampled queries conditional on recorded ratings; they do not measure inter-rater uncertainty. As these ratings set the four calibration means, annotation differences could propagate to scenario activation. Section VII specifies a double-rating audit and a check of its effects on the anchors. These prospective analyses have not been conducted.

The corpus lacks search-surface metadata, locale, timestamped snapshots, archive URIs, and content hashes. Five queries per vertical cannot support prevalence estimates or exact retrieval replication. Scenario rows contain no live rankings, real publishers, automated reviewer calls, or human query labels. Router metrics use the scenario activation rule and one family split; split-to-split variation is unknown. No learned page classifier, matched RAG generation, answer ratings, or serving latency are evaluated. Conclusions remain limited to pilot score behavior, workload under the generator, controlled-reformulation routing, and oracle risk–coverage analysis.

## VII. Auditability and Claim Gate

Table IV defines the claim boundary. The analysis retains raw and calibrated scores, seed 20260716, and primary threshold $\tau = 0.50$. Unresolved rows remain missing; stored geometric means agree within 0.00046 of rounded values. Table V summarizes the next-stage tests. The following protocols are prospective, not additional results.

TABLE IV. Claim-to-Evidence Gate

| Result | Supported claim | Unsupported claim |
|---|---|---|
| Manual anchors | Calibration targets | Reliability or prevalence |
| Scenario | Behavior under generator | Live-Web rate |
| Router test | Controlled mutation recovery | Deployed-query accuracy |
| Oracle gate | Risk–coverage target | Learned classifier effect |
| RAG protocol | Evaluation protocol | Causal answer improvement |

Runtime records should preserve query/family identifiers, URL, rank, evidence, dimension scores, uncertainty, and policy action. Failed access requires review or replacement, never an automatic pass.

TABLE V. REQUIRED NEXT-STAGE EVALUATION

| Component | Held-out unit / outcome | Utility constraint |
|---|---|---|
| Annotation | Overlap / weighted agreement | Evidence sufficiency |
| Query router | Family / macro-F1 and recall | Activation workload |
| Page classifier | Family/domain / action P/R | Abstention and access |
| RAG answer | Paired query / disclosure | Support and latency |

For each policy action, the record should identify the supporting passage or ownership evidence and preserve the original rank. This would let a later audit trace an activation decision back to the pages that contributed to the family score. Recording retained and withheld pages together would also expose cases where exclusion removes the only available support for a claim.

An overlap audit would use 40 resolved pages, ten per vertical, spanning reviewers, queries, and both sides of 0.50 where available. Two raters independently score identical archived evidence with a frozen codebook, blinded to prior and each other's ratings. Before adjudication, report linearly weighted Cohen's κ per dimension, unweighted κ at the 0.50 composite threshold, raw agreement, and query-cluster bootstrap intervals. Report stratum counts and unresolved pages separately; then check adjudication's effects on anchors and activation.

The overlap audit should retain both original ratings before adjudication. Reporting disagreement by dimension would show whether uncertainty concerns the source relationship, disclosure, or apparent independence. Recomputing the anchors from the paired ratings would then test how that uncertainty changes family activation under the same frozen scenario, without treating adjudicated scores as independent validation.

For retrieval, archive ranked URLs and content for new natural queries in all four verticals. Record retriever settings, locale, time, ranks, publishers, access status, and hashes. Fix calibration on disjoint development data. On held-out families, compare activation, retention, and publisher diversity at matched top-k; repeat snapshots to assess temporal change. Report each vertical with query-cluster intervals and interpret transfer only for that retrieval setting.

The archived-retrieval comparison should report the number of candidate pages available before each policy action. Keeping those candidates fixed would separate the effect of the triage rule from changes in retrieval. Comparisons across snapshots would then address temporal instability, while reporting publisher diversity alongside retained evidence would make concentration visible.

## VIII. CONCLUSION

Query-aware triage separates family activation from page treatment. The single-coded pilot supplies provisional anchors; the calibrated 20,000-row synthetic scenario supports controlled aggregation, reformulation, routing, and workload analysis. The oracle curve defines a risk–coverage target, not a measured classifier benefit. Aspect-attentive sponsorship detection motivates a later page model [15], without validating it here. Reliability and transfer to real retrieval remain open; we specify double-rating and archived-retrieval protocols to address them. Present activation rates are conditional on the generator and do not estimate deployed review workload, live-Web prevalence, or downstream answer-quality gains.

## ACKNOWLEDGMENT

This research received no external funding. The authors declare no conflicts of interest. GPT-5.6 was used to polish the language and correct grammar.